Harry Rodger[a,b] and Andrew Lensen[a*] and Marcin Betkier[b]

[a]*School of Engineering and Computer Science, Victoria University of Wellington, Wellington, New Zealand;*
[b]*School of Law, Victoria University of Wellington, Wellington, New Zealand*

# Explainable Artificial Intelligence for Assault Sentence Prediction in New Zealand[1]

**Abstract**

**The judiciary has historically been conservative in its use of Artificial Intelligence, but recent advances in machine learning have prompted scholars to reconsider such use in tasks like sentence prediction. This paper investigates by experimentation the potential use of explainable artificial intelligence for predicting imprisonment sentences in assault cases in New Zealand's courts. We propose a proof-of-concept explainable model and verify in practice that it is fit for purpose, with predicted sentences accurate to within one year. We further analyse the model to understand the most influential phrases in sentence length prediction. We conclude the paper with an evaluative discussion of the future benefits and risks of different ways of using such an AI model in New Zealand's courts.**

## Introduction

Artificial intelligence (AI) has been widely used in a range of interdisciplinary tasks in recent years, with success in protein folding (Jumper et al. 2021), game playing (Silver et al. 2018), and text generation (Brown et al. 2020), making headlines throughout the popular press. AI could also potentially undertake some tasks in criminal justice system, which is a topic of a fierce debate among scholars and practitioners. The use of the COMPAS algorithm calculating the risk of reoffending in the US courts has been challenged in the famous *Loomis v. Wisconsin* case (*Loomis v. Wisconsin* 881 N.W.2d 749 (Wis. 2016)), in which the court rejected the appeal against the decision supported by the COMPAS, but also warned about the risks of using such algorithms and their transparency. Later, ProPublica published a study showing that the algorithm is biased

---

against black people (Angwin et al., 2016) which was rebutted by the company providing the algorithm (Dietrich et al., 2016), and challenged as erroneous and unethical (Flores at al., 2016).

Despite that heated debate, the further advances in the use of machine learning (ML) for text/document decision-making ("natural language processing (NLP)") has prompted AI and legal scholars to reconsider different uses of ML in the sentencing process(Stobbs et al. 2017; Re and Solow-Niedermann, 2019; Donohue, 2019). Also, the courts in different countries actively investigate the use of AI in different stages of criminal proceeding. For example, in China, AI 'service robots' answer questions about the case law, provide legal consultations and predict the risk and likely outcome of the lawsuit to help potential litigants (Wang, Tian, 2021, pp.211-2). Further, they help judges to adjudicate small claims and disputes showing recommended sentence based on a list of input factors (ibid, also Stern et al. 2021, 526-7). Similarly, in two states of Malaysia, a common law country, the AI is currently being tested as producing judgment recommendations for two serious criminal offences: drug possession and rape (Lim and Gong, 2020, also 'Mr. Robot…' *Daily Sabah*, 12 April 2022).

In this paper, we explore the potential application of ML for predicting sentencing in criminal assault cases in the courts of Aotearoa–New Zealand. We begin with an overview of sentencing in New Zealand, focusing on sentencing determination in assault cases. We then introduce a proof-of-concept AI model that predicts sentence length in assault cases based on 302 decisions of the New Zealand courts. We evaluate the model's accuracy and investigate its "explainability" to understand the most relevant phrases it considers when predicting a sentence. We conclude the paper by discussing the results of our experiment: showing potential benefits and risks of different applications of AI predictions in the sentencing process.

To the best of our knowledge, this is the first practical investigation of applying AI to this task in a New Zealand context.

## Sentencing in New Zealand

Sentencing is the process of deciding on the type and length of a criminal sentence. It is one of the last elements of the criminal process after the relevant facts have been established and the defendant has been found guilty of criminal conduct. Finding an adequate sentence for a particular offence is a complex task. The "algorithm" to make a sentencing decision is partially included in the Sentencing Act 2002. That Act provides a framework for criminal sentencing and lists its purposes (s7), principles (s8) and aggravating and mitigating factors (s9). Those factors implicate imposing a harsher or, accordingly, more lenient sentence. However, the existence of the statutory framework and the list of factors used by the judges as indicators does not provide a complete algorithm for the judge nor necessarily make the sentencing task simpler (e.g., Roberts 2003, p.19).

Sentencing is complex because it may need to fulfil many goals of the justice system, which are often diverging. For example, according to different punishment theories, the criminal sentence should deter from committing a crime, incapacitate the offender, provide retribution for their wrongdoings, and/or allow that person to return to society as a rehabilitated member. (Donohue 2019; also, s 7 Sentencing Act 2002) Considering those different goals, it is no surprise that sentences delivered by judges may differ even for apparently similar cases.

The New Zealand sentencing system is based in parallel on two premises that may be seen as contradictory – judicial discretion and consistency (Law Commission 2006 R94, pp.17-20). Judicial discretion means that the judge can freely decide within the law's limits. It enables the judge to individualise the sentence after a complete evaluation of the factors affecting the circumstances of each case (*Hessel v R* [2010] NZSC 135, para. 27). Judges can do this because of their experience in applying the law, understanding societal norms and human conditions, and intuition and sense of justice.

The downside of the discretionary character of sentencing is that it causes inconsistencies between the decisions (Law Commission 2006, p. 9; in criminology research, e.g., Goodall 2014; also "Judges' sentencing lacks consistency - Andrew Little" Newshub 2017).  Those inconsistencies may arise precisely because of the individualisation of the judgments, but they may also appear because of considering some extraneous factors. For example, the 2011 research on decision making in Israeli courts found that the percentage of favourable decisions drops from 65% to nearly 0% before the food break and returns to 65% after such a break (Danziger et al., 2011). Even though that was later partially explained as the effect of rational time management (Glöckner, 2016), it was widely published worldwide and suggested that hunger influences judicial decisions.

This example clearly shows not only a need to exclude irrelevant factors from sentencing but also a need to clearly demonstrate consistency in decision making both to the offender and the public. It is necessary to maintain the confidence that similar offences are treated similarly (s 8(e) of the Sentencing Act 2002). Further, that consistency should be maintained between the decisions of the same judge and across the decisions of different judges in different court locations across the entire country. That is because a justice system that delivers inconsistent and uneven results risks losing the trust of the public and its cooperation (See, e.g., Mallett 2015, p. 535 ff.).

The need for consistency creates a trend towards a further 'algorithmising' of sentencing, quantifying the decision into smaller elements and mandating the results. In this respect, the most critical measure is guideline judgments of higher-level courts. Guideline judgments specify the decision-making process relating to a particular type of offending in more detail. Historically, the Court of Appeal presiding over appeals in criminal cases developed in its jurisprudence the principle of consistency which was used to review the sentences of the lower courts (*Hessell v R*, para. 24–25; Holt, p. 5). That practice, reinforced by the parallel developments in the English jurisdiction, was transformed into issuing judgments that, apart from deciding the case in question, also explicitly provide guidelines to the lower court judges regarding the treatment of similar cases (Holt, p. 7–11). Since the enactment of the Sentencing Act 2002, that practice has continued in New Zealand.

There was also a further attempt towards 'algorithmising' the sentencing by a dedicated body, the Sentencing Council, that could issue sentencing guidelines (Law Commission 2006). The Parliament enacted the Sentencing Council Act 2007, providing necessary laws. Despite that, the Council was never established because the new government (coming from a different political party) had significant objections related to the fundamental constitutional rules, such as judicial independence and separation of powers (Mallett 2015, p. 561–566). A few years later, a similar critique was presented in *Hessell* by the judges of the Supreme Court (*Hessel v R*, para. 67). As a result of those setbacks, sentencing in New Zealand is still based on guideline judgements that still experience the same consistency problems as diagnosed by the Law Commission in 2006 (Law Commission, 2006, p.20).

### Sentencing for assault

Assault is defined in the Crimes Act 1961 in section 2(1) as "the act of intentionally applying or attempting to apply force to another person" either directly or indirectly. Assault can be delineated into different categories, which carry varying penal consequences depending on their seriousness:

(1) Wounding with intent to cause grievous bodily harm (GBH, s 188(1)). It can give rise to a sentence of 14 years imprisonment. The Court of Appeal issued a guideline judgment for that offence in *R v Taueki* (*R v Taueki* [2005] NZCA 174). That decision could also be applied to other offences, especially to aggravated wounding (s 191(1)).

(2) Wounding with intent to injure. It is essentially the same offence as s 188(1), but with a different mental element - intent to injure. The relevant guideline judgment for that offence is *Nuku v R* (*Nuku v R* [2012] NZCA 584), which expands on *Taueki* and modifies some of its guidelines (due to the lower statutory maximum sentence of 7 years).

(3) Assault in the form of strangulation or suffocation (s 189A). The relevant guideline was defined in a High Court decision *Ackland v Police* (*Ackland v Police* [2019] NZHC 312).

(4) A common assault only has a maximum term of imprisonment of one year.

Guideline judgments give a two or three-step algorithm in which the first step relies on setting a 'starting point' for the sentence. Setting up the starting point is done by considering the factors related to the offending indicated in the guideline judgment, identifying them (e.g., violence, cruelty, premeditation, provocation) and evaluating them (e.g., how much premeditation) (*R v Taueki*, para. 26–33). Many of those factors are defined in s 9 of the Sentencing Act 2002. Based on those considerations, the Court identifies the bands within which the starting point should be located. For example, in *Taueki,* the court identified three such bands: 3-6 years, 5-10 years and 9-14 years of imprisonment. With no aggravating factors present, the starting point will be in the first band, with 2-3 aggravating factors in the second band, and so on. It is worth noting that the court reserves itself a broad discretion within those bands and emphasises evaluation and avoiding "a danger of formulaic or mathematical approach" to sentencing (*R v Taueki*, para. 30).

When the starting point is set, the court considers the offender's particular personal circumstances, which may lower or increase the starting point. Early guilty plea, assistance to authorities, age, and other factors contribute to setting the percentage of decrease or increase of the final sentence (*R v Mako* [2000] NZCA 407, para. 62). For more serious offendings (like GBH), the court will consider the minimum period of imprisonment that the offender will be required to serve according to s 86 of Sentencing Act 2002 (*R v Taueki*, para. 47–58).

Thus, the sentencing decision contains both the description of the facts of particular offending and judicial analysis based on the relevant sentencing judgments, which lead to the determination of the sentence. The input of the AI model will have to consider all those elements.

**An AI model for assault sentence prediction**

Machine learning (ML) (a category of AI) has increasingly been applied to various diverse, real-world applications. ML algorithms "learn from examples" by analysing a *training set* to learn patterns in data that relate to the *target* goal. In this context, our training set comprises a set of court cases, with the target being the sentence length of each case. By using this training set, the ML algorithm can learn a *model* that takes a case and produces a sentence. In addition to the training set, a *test set* is also used in machine learning. The test set – also called "unseen data" – is **not** used by the ML algorithm to learn (train) the model. Instead, it provides an independent evaluation of the quality of the learned model at the end of the training process. The results on the test set are a measure of how well the model is expected to perform in practice on data it has not been exposed to previously. A third set – the *validation set* – is often also used in ML. This validation set is used during the design and training of the ML algorithm, to evaluate the performance of the algorithm under different hyper-parameter settings and to perform "early stopping" in the training process. In this paper, we use a train/validation/test split of 65%/10%/25%, where the data is first randomly shuffled to avoid any chronological patterns. The test set is only used to generate the final results – it does not influence any other decisions regarding the overall approach taking.

*Dataset Creation*

We collected 302 unique assault decisions from the New Zealand Legal Information Institute (NZLII) database (http://www.nzlii.org/), with sentence lengths between zero and 14.5 years. Cases either had assault as the lead charge or an interrelated offence (such as aggravated robbery, sexual assault, grievous bodily injury). For example, aggravated robbery is an interrelated offence as it requires an aggravating act by the perpetrator (e.g., an assault) during a robbery. Each case was manually labelled by the number of months of the sentence handed down.

Each case was converted from PDF to text format, resulting in a set of text documents, each with between 436 and 6196 words. These documents were further pre-processed to remove sentence lengths (i.e. removing the class information), tabs, newlines, and null characters. The phrases "home detention" ,"community detention", and "preventative detention" were also removed to reduce information leakage. We also removed all months and year words from the text, punctuation (including accents), and stop words ("useless" words such as "the"," a", "an", ...).

### *Feature Engineering*

We use an *n-gram* model (Manning and Schutze 1999) to represent each document – a common approach in NLP and computational linguistics. An *n*-gram represents a phrase of length *n* in a piece of text. For example, a 1-gram (unigram) could be the word "assault"; a 2-gram (bigram) would be "aggravated assault", and a 3-gram (trigram) would be "aggravated assault with". In this model, each document is represented as a *vector* (list of values) of n-grams, specifying the number of occurrences ("counts") of each n-gram in that document. For example, "aggravated assault" may occur twice in a document, giving a vector value of 2 – or it may never occur, giving a vector value of 0.

We use n-grams up to an *n* of 3 in this work (i.e. unigrams, bigrams, and trigrams), as higher-order n-grams give very large vectors and did not improve accuracy. To exclude overly specific or broad phrases, we do not include n-grams that occur in fewer than three documents or more than 90% of the documents.

Using raw n-gram counts is known to overly emphasise the importance of more common words – for example, if a word occurs only in ten documents, it is unlikely to be found "important" by the AI model, even if it is strongly characteristic of those ten cases. To remedy this, most work in NLP uses a transformation such as tf-idf (term frequency-inverse document frequency), which considers both how frequently an n-gram appears in a document and the total number of documents (in the training set) that contain that word. This measure is calculated as:

$$tfidf(t, d) = tf(t, d) \times idf(t)$$

$$tf(t, d) = \frac{f_{t,d}}{\sum_{t' \in d} f_{t',d}}$$

$$idf(t) = \log \frac{1 + n}{1 + df(t)} + 1$$

where *t* is a given n-gram, *d* is a document, $f_{t,d}$ is the number of occurrences of *t* in *d*, $df(t)$ is the number of documents containing at least one occurrence of *t*, and *n* is the total number of documents in the training set.

The final representation of each document is a tf-idf vector of n-grams. This is a *sparse* representation, as many n-grams will not occur in a given document. This representation is also advantageous for model interpretability (e.g. in contrast to an embedding approach), as each "feature" in the vector is directly derived from a specific n-gram.

Stemming is a process in NLP that reduces words to their base "stem". For example, applying stemming to "sentence" and "sentenced" would reduce both words to "sentenc". This is often helpful for ensuring words with the same meaning are treated the same by a ML algorithm. We do not use stemming in this work due to the importance of tense and plurality information in case summaries. Consider, for example, "victim" vs "victims": both stem to "victim", but the second word is indicative of a multiple assault and a potentially longer sentence.

### The AI Model

The prediction of sentence length can be framed as a *regression* problem, where the AI model learns to predict the sentence in number of months. In this exploratory study, we want to use an *explainable* model: we are particularly interested in what phrases the model weights most heavily when calculating a sentence. To achieve this, we utilise a linear regression algorithm, which searches for the best set of weights for the set of n-grams. Specifically, we use:

- Stochastic Gradient Descent (SGD) as our optimiser (SGD is well-suited to sparse data such as text);
- An $\epsilon$-insensitive variant of squared error as our loss function, which treats small errors as zero[2] with with $\epsilon$=0.1. This helps to reduce overfitting during training. ;
- L1 (Lasso) regularisation with an alpha=0.001 to encourage feature selection (i.e. using as few phrases as possible in the model);
- A maximum of 2000 iterations of gradient descent; in practice, the search converges much more quickly due to the use of early stopping.

A training/test split of 75/25%, with 10% of the training set used as a validation set for early stopping. We note that more advanced algorithms could be used (e.g. polynomial regression, tree-based methods, neural networks), but considering the exploratory nature of this paper, we purposefully choose a simple linear regression approach that is innately straightforward to interpret. We expect to explore more complex explainable approaches in the future.

### Results

After training, the model had a coefficient of determination ($R^2$) of 0.524 on the test set (unseen data), as shown in Figure 1, which corresponds to a moderately strong correlation.

---

[2] Defined as: $loss = \max(0, |y - p| - \epsilon)^2$ for known $y$ and predicted output $p$.

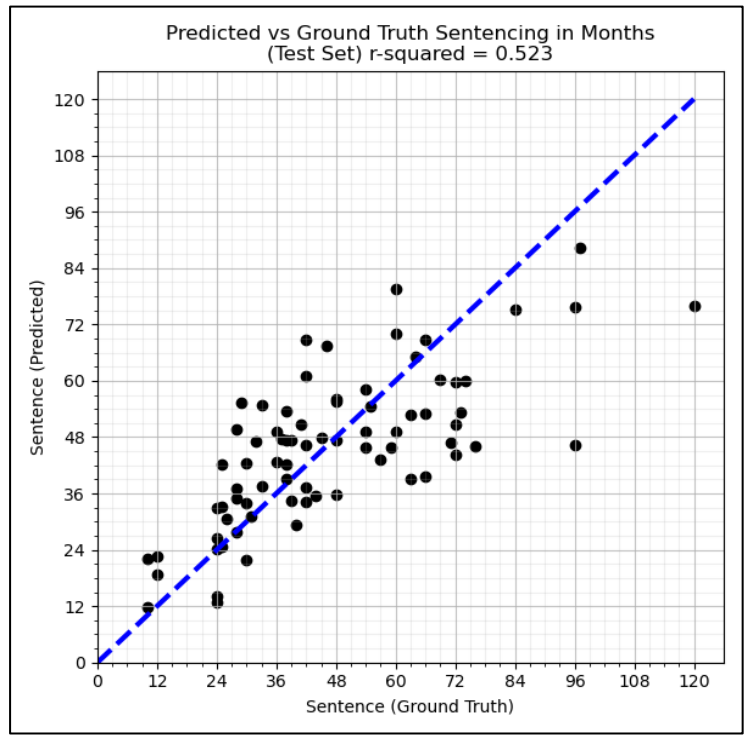

*Figure 1: Predicted sentence vs correct sentence ("ground truth")*

The mean absolute error (MAE) on the test set was 11.76, i.e., on average, the predicted sentence of a case is incorrect by 11.76 months. It is worth noting that 'incorrect' means here 'different from the real judgment in particular case', as we do not know what sentence would be 'correct'. Considering that sentences range from 0 to over ten years, an error of under one year for a first exploratory study is encouraging. These results are lower than on the training set (MAE of 6.05 and $R^2$ of 0.89), which is expected in machine learning – but this gap could likely be closed with further research. Having said that, it is impossible to expect error value closing to 0, because we operate on the set of decisions that are already inconsistent. That is, they are individualised by the judges and there are no human experts which evaluate them further (excluding the cases appealed to the higher courts). Achieving the low level of error may only be a starting point to a discussion into reasons for particular errors (or types of errors).

### Interpretation of the trained model

The phrases (n-grams) assigned the largest positive weights by the model are shown in descending order in Figure 2. A phrase with a darker colour has a more significant weighting (e.g. "sexual" has a weighting of 116.75), meaning that its presence in a case summary will cause the AI model to predict a significant **increase** in the sentence[3]. A phrase with a longer bar is present in a higher proportion of the documents (court cases), affecting the sentence prediction more frequently. Figure 3 shows the phrases with the most significant **negative** weights, i.e., those associated with the largest **decrease** in sentence predictions. When computing the weights, we take the document frequency into account by multiplying the model's weight by the IDF of the n-gram. This ensures that the commonality of a n-gram does not affect its position in our rankings.

---

[3] Note that we use "cause" here to refer to the phrases that the AI model associates with sentence predictions. We do not suggest that there is a specific causality between a word occurring and a human judge making a certain decision.

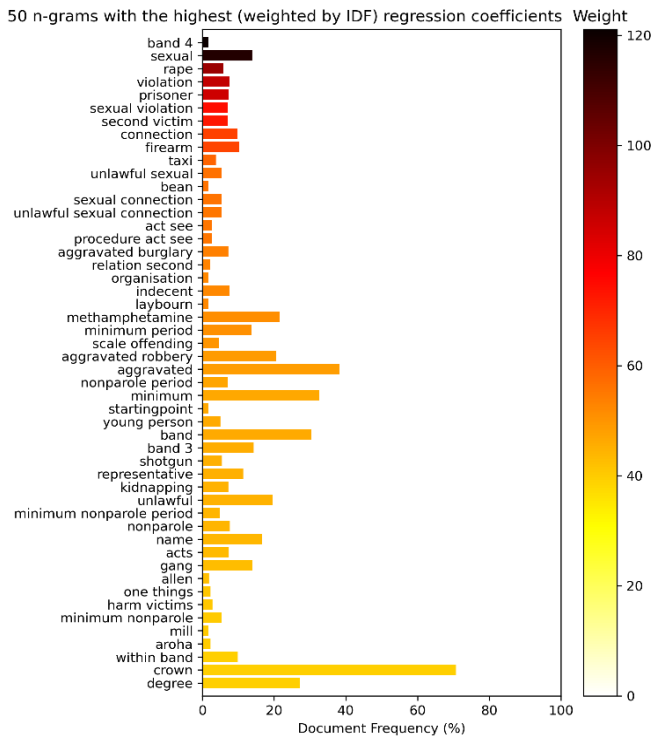

*Figure 2: Phrases indicative of longer sentences*

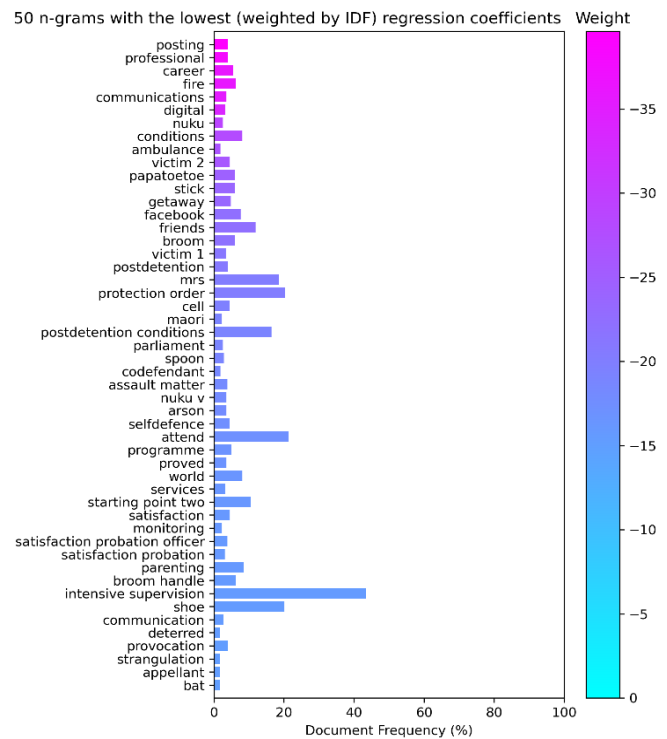

*Figure 2: Phrases indicative of shorter sentences*

We discuss a selection of phrases of particular interest below:

(1)  "**sexual**" has the second-largest positive weight in the model (corresponding to a higher sentence). This word may be linked with an indecent assault charge or other sexual crimes. Maximum sentences for sexual crimes are higher than common assault, at seven years compared to one year. Where the phrase "sexual" appears multiple times in a document, it will add to the sentence cumulatively. This is consistent with cases where there are multiple independent charges in a single sentencing, or where the judge refers to a detail of the offending multiple times to reinforce their decision-making.

(2) **"Communications"** and **"Digital"** refers to the Harmful Digital Communications Act 2015 (HDC Act). The maximum imprisonment sentence which can be given to a person charged under the HDC Act is 2 years. The model correctly associates "Communications" and "Digital" with a lesser sentence than other lead charges such as GBH (which has a maximum sentence of 14 years under the Crimes Act 1961).

(3) **"band 3/4"** refers to various bands for the 'starting point' of a sentence (based on guideline judgements). A higher band indicates a higher starting point and thus a higher sentence. The model correctly associates "band 4" with a higher weighting than "band 3".

(4) "**Nuku**" refers to the guidance judgment *Nuku v R,* which applies for wounding with intent to injure – a crime associated with a less severe sentence (up to 7 years) than wounding with intent to cause GBH. The model correctly associates it with a reduction in sentence length.

(5) The "**Crown**" " is used to refer to the prosecution counsel in criminal proceedings. The model may be using its presence to shift the intercept of the linear regression model towards the average sentence length (to minimise prediction error). We also suspect that more serious cases may reference more often to the Crowns arguments in favour of a higher sentence, which correlates to more frequent use of this phrase.

Sentencing decisions contain both the description of a particular case and the procedural description of the sentencing method, such as references to guideline judgments. The AI model can correctly capture the relationship between procedural elements and sentence lengths; procedural elements related to aggravated crimes are associated with longer sentences.

If we exclude these procedural elements, we are left with phrases relating to case facts only. Many of them are intuitive, for example, "violation", "rape", "victims", "methamphetamine", and "firearm" for longer sentences. However, some of these phrases are harder to explain – it is not immediately apparent why "taxi"is associated with a longer sentence and "fire", "ambulance", or "spoon" with shorter sentences. There may be some amount of overfitting (the model learning the specifics of individual cases rather than general patterns) or other factors – we intend to investigate this further in future work.

**Applications of the Model**

The AI sentencing algorithm could provide its users, researchers, or practitioners, with a range of applications. A selection of potential applications is briefly discussed below.

*Automating the sentencing process*

The AI model could potentially be presented with the facts of a new criminal case, with it indicating an indicative prison sentence. Assuming that the mean prediction error could be further decreased, that could potentially be used to automate the sentencing process for some crimes. This may increase consistency across sentencing in the New Zealand criminal justice system. Further, the time it takes for a judge to give a sentence could be shortened due to the automatic prediction that the model would deduce. This, in turn, could reduce the cost for the state to fund sentencing and for the parties to participate in lengthy litigation. Also, that gives hope for clearing the backlogs in courts and reduce the number of people on remand, which is increasing ('Projected growth…', Newsroom, 04/04/2022).

While AI making decisions about the length of prison sentences may feel uncomfortable, we note that in many areas of law enforcement, decision making of the penalty is already governed by algorithms. Those algorithms are usually defined by law and are – to some extent – automated. For example, the penalties for parking and speeding offences are precisely defined in the Land Transport (Offences and Penalties) Regulations 1999. Those penalties are fully quantified, eliminating discretion and allowing perfect consistency and transparency. An AI algorithm trained on previous sentencing decisions may be seen as just another algorithm.

However, operating an AI algorithm is notably different from the way the 'legal algorithm' is used, and the importance of the sentence for a person is different. Further, the judges in a common law system do not only apply the law, but also change or adapt the law and, importantly, deliver individualised justice that should be open to scrutiny and include societal and equitable interests (Campbell, 2020, p. 339 ff.). An AI algorithm does not readily provide full explainability and transparency of the decision-making process, and incorporating all previous sentencing decisions could reinforce existing biases (e.g., Gavaghan et al. 2019). Also, the automation of the process could prevent judges from individualising justice by considering exceptional factors. We cannot expect AI to be compassionate, merciful and make moral decisions; also, we are surely not ready for AI changing the law, for example, when a particular case is unprecedented. These drawbacks are precisely the 'danger of formulaic or mathematical approach' described by the Supreme Court in *Hessell* (*R v Hessell* [2010] NZSC 135 para.29). So, AI algorithms should not be automated because there must be room for a judge to act discretionarily, which will inevitably introduce some inconsistency in decisions. That 'desirable inconsistency' should be distinguished from 'undesirable inconsistency', like bias, or unwanted extraneous factors, like hunger in the example above.

### Tool for judges to improve sentencing

Despite the limitations described above, there is still potential for AI to be used as a tool for judges. The AI model incorporates the knowledge from all previous sentencing decisions and analyses them lexically. Firstly, it could be used to verify and improve sentencing by, for example, delivering predictions before making the final decision or refining guideline judgments. The model's partial explainability could help judges determine and evaluate the factors related to specific offending considering the analysis of the previous cases. It would also extend the current computer systems' ability to provide a database of decisions (cf. Donoghue 2019, pp. 673-675). It is also possible to benefit from an AI approach even without performing prediction; using an unsupervised approach, where the judge is instead shown *similar* historical cases (and their accompanying sentences) would allow the judiciary to much more easily cross-reference their decision-making with prior cases.

Secondly, it could help the judges reassess the guideline judgments by tracking their use and re-evaluating the aggravating and mitigating factors. The model could help scrutinise how the sentencing judgments are applied in practice – AI could augment judges' abilities instead of substituting them (Donohue 2019, pp. 671-672). Also, unlike the previously proposed Sentencing Council, this approach would not undermine judicial independence and judicial discretion as the tool would be entirely in the hands of judges.

There are also potential disadvantages of such AI application, which are very similar to the ones outlined in the discussion around COMPAS algorithm: the opacity and incomplete explainability of the algorithm, the risk of bias in the training data set (because of the method of collection or, importantly in New Zealand, the entrenched pattern of historical bias) or the risk of 'automation bias', so the outcome in which judges would be willing to rely on the AI recommendation without analysing it further (Campbell, 2020 pp. 342-3; Zalnieriute, 2021; Spiekermann, 2008, pp. 32-4).

### *Tool for lawyers*

Lawyers may benefit from utilising the prediction functionality of the model with a particular assault case. The AI's prediction may better inform them of what to expect in terms of the sentence. They may have some estimation about the case outcome which may be helpful to decide about the plea. Furthermore, the lexical analysis of the sentence of a particular judge may give them a list of terms that correlate with decisions about higher or lower sentences. That, in turn, could inform about the way the case should be argued to direct it towards lower (or higher) sentence.

### *Research tool to analyse the justice system*

Finally, this type of model could help researchers of the criminal justice system in New Zealand. It could allow them to analyse sentencing better, initiate critical discussions, provide new arguments, and empower or disprove existing ones. For example, such a tool could help to demonstrate the factors for judicial decision-making and perhaps show whether some extraneous factors, like hunger, have been taken into account. However, it is paramount that researchers understand limitations such as potential bias or overfitting.

### Future Work

There are many possible directions of further work to this preliminary study, such as:

- further investigation of the use of specific n-grams in the model. For example, why are 'relation' or 'young' are indicators of longer sentences?
- further development of a system for explaining specific decisions to users. Such a system could show the contribution of both higher-order and lower-order n-grams (e.g. "sexual assault", "sexual", and "assault") to the prediction.

- inclusion of more decisions as input to the model and checking how that influences the prediction error and the distribution of n-grams.
- a comparative analysis of different main centres of the District Court, court locations or even particular judges. That could uncover any existence of the lack of consistency between different locations or judges.
- analysing counterfactual scenarios, where the model is provided slightly different cases that differ in specific ways, such as the offender's ethnicity. That could help explore whether the sentences correlate with factors that should be irrelevant and discriminatory.
- analysing the application of a particular guideline judgment (e.g., *R v Taueki*) to reflect on its suitability and consistency of use.
- exploring other types of offences.

## Conclusion

The sentencing process in New Zealand's criminal justice system already embraces rudimentary algorithms described in the statutory law and issued by the courts in the form of sentencing guidelines, which inform a judge's sentence. Perhaps, the use of an Artificial Intelligence algorithm is nothing to be afraid of – it would be simply a more complex algorithm than what already exists.

The developed AI model allows predicting sentences for assault with an average error of 11.75 months. It also allows investigating specific phrases (n-grams) that cause the model to drive the predictions up or down. The limited scope of the research does not allow for using that model in practice, but it provides a valuable proof-of-concept in exploring the capabilities and uses of such a system. We also expect that further exploration into more sophisticated explainable regression algorithms in the future (such as polynomial or tree-based approaches) could improve model accuracy with only minor impacts on the explainability of the model.

There are still concerns about the transparency of such predictions and the full explainability of the model, preventing its use for automating sentencing. Also, such use could hinder judges in fully individualising sentences. However, such a model could help judges better exercise judicial discretion to deliver more consistent results by giving them a tool to verify and analyse sentences or guideline judgments. Such a model could ultimately be an invaluable help in future sentencing.

## Acknowledgements

The authors would like to thank Professor Yvette Tinsley from Victoria's Law School for her very valuable remarks. Also, this work would be impossible without the availability of court decisions through New Zealand Legal Information Institute. We are grateful for the NZLII's huge contribution to the availability of legal information for research and study in New Zealand.

**References**

*Ackland v Police*. 2019. [place unknown]; [accessed 2021 Sep 18].
http://www.nzlii.org/nz/cases/NZHC/2019/312.html

Brown T, Mann B, Ryder N, Subbiah M, Kaplan JD, Dhariwal P, Neelakantan A,
Shyam P, Sastry G, Askell A, et al. 2020. Language models are few-shot
learners. In: Larochelle H, Ranzato M, Hadsell R, Balcan MF, Lin H, editors.
Advances in neural information processing systems [Internet]. Vol. 33. [place
unknown]: Curran Associates, Inc.; p. 1877–1901.
https://proceedings.neurips.cc/paper/2020/file/1457c0d6bfcb4967418bfb8ac142f
64a-Paper.pdf

Devlin J, Chang M-W, Lee K, Toutanova K. 2019. BERT: Pre-training of Deep
Bidirectional Transformers for Language Understanding. In: Proceedings of the
2019 Conference of the North American Chapter of the Association for
Computational Linguistics: Human Language Technologies, Volume 1 (Long
and Short Papers) [Internet]. Minneapolis, Minnesota: Association for
Computational Linguistics; [accessed 2021 Dec 8]; p. 4171–4186.
https://doi.org/10/ggwf6Michael E. *A Replacement for Justitia's Scales?:
Machine Learning's Role in Sentencing* (2019) Harvard Journal of Law &
Technology Volume 32, Number 2

Flores, Anthony W, Christopher T Lowenkamp, and Kristin Bechtel. "False Positives,
False Negatives, and False Analyses: A Rejoinder to 'Machine Bias: There's
Software Used Across the Country to Predict Future Criminals. And It's Biased
Against Blacks.'" *Federal Probation Journal* 80, no. 2 (September 2016):
38.Goodall W. 2014. Sentencing consistency in the New Zealand District Courts
[accessed 2021 Sep 15]. http://researcharchive.vuw.ac.nz/handle/10063/3375

Holt S. Appellate sentencing guidance in New Zealand. NZPGLEJ  [accessed 2021 Sep
15] (3).
https://cdn.auckland.ac.nz/assets/nzpglejournal/Subscribe/Documents/2006-
1/2_saul.pdf

Jumper J, Evans R, Pritzel A, Green T, Figurnov M, Ronneberger O, Tunyasuvunakool
K, Bates R, Žídek A, Potapenko A, et al. 2021. Highly accurate protein structure
prediction with AlphaFold. Nature. 596(7873):583–589.
https://doi.org/10/gk7nfp

Law Commission. 2006. Sentencing guidelines and parole reform  Wellington; [accessed 2021 Sep 15]. https://www.lawcom.govt.nz/sites/default/files/projectAvailableFormats/NZLC%20R94.pdf

Lim, Claire, and Rachel Gong. "Artificial Intelligence in the Courts: AI Sentencing in Sabah and Sarawak." *Khazanah Research Institute*, August 18, 2020, 10.

Mallett SJ. 2015. Judicial discretion in sentencing: a justice system that is no longer just? :40.

Manning C, Schutze H. 1999. Foundations of Statistical Natural Language Processing. MIT Press.

Re, Richard M. & Solow-Niederman, Alicia *Developing Artificially Intelligent Justice (2019) 22 Stanford Technology Law Review 242*

Roberts JV. 2003. Sentencing Reform in New Zealand: An Analysis of the Sentencing Act 2002. Australian & New Zealand Journal of Criminology. 36(3):249–271. https://doi.org/10.1375/acri.36.3.249

Silver D, Hubert T, Schrittwieser J, Antonoglou I, Lai M, Guez A, Lanctot M, Sifre L, Kumaran D, Graepel T, et al. 2018. A general reinforcement learning algorithm that masters chess, shogi, and Go through self-play. Science. 362(6419):1140–1144. https://doi.org/10/cxq3

Spiekermann, Sarah. *User Control in Ubiquitous Computing: Design Alternatives and User Acceptance*. Aachen: Shaker. 2008.

Stern, Rachel E, Benjamin L Liebman, Margaret Roberts, and Alice Z Wang. "Automating Fairness? Artificial Intelligence in the Chinese Court." *Columbia Journal of Transnational Law*, n.d., 40.

Stobbs, Nigel, Hunter, Dan and Bagaric, Mirko *Can sentencing be enhanced by the use of artificial intelligence?* (2017) 41(5) Criminal Law Review 261

Vaswani A, Shazeer N, Parmar N, Uszkoreit J, Jones L, Gomez AN, Kaiser Ł, Polosukhin I. 2017. Attention is All you Need. In: Advances in Neural Information Processing Systems Vol. 30.: Curran Associates, Inc.; [accessed 2021 Dec 8]. https://papers.nips.cc/paper/2017/hash/3f5ee243547dee91fbd053c1c4a845aa-Abstract.html

Wang, Nyu, and Michael Yuan Tian. "'Intelligent Justice': AI Implementations in China's Legal Systems." In *Artificial Intelligence and Its Discontents: Critiques* *from the Social Sciences and Humanities*, edited by Ariane Hanemaayer, 197–222. Cham: Springer International Publishing. 2022.

Zalnieriute, Monika. "Technology and the Courts: Artificial Intelligence and Judicial Impartiality." *SSRN Scholarly Paper*, June 16, 2021. https://papers.ssrn.com/abstract=3867901.

*R v Taueki* [2005] NZCA 174

*Nuku v R* [2012] NZCA 584

*Hessel v R* [2010] NZSC 135

*R v Mako* [2000] NZCA 407

*Ackland v Police* [2019] NZHC 312

*Loomis v. Wisconsin* 881 N.W.2d 749 (Wis. 2016)

Judges' sentencing lacks consistency - Andrew Little. 2017. Newshub [accessed 2021 Sep 15]. https://www.newshub.co.nz/home/politics/2017/12/judges-sentencing-lacks-consistency-andrew-little.html

Julia Angwin, Jeff Larson, Surya Mattu, Lauren Kirchner *Machine Bias: There's Software Used Across the Country to Predict Future Criminals. And it's Biased Against Blacks.* ProPublica *2016-05-23 https://www.propublica.org/article/machine-bias-risk-assessments-in-criminal-sentencing Accessed 2021-12-10*

Anadolu Agency. "Mr. Robot Takes on Law&order: Malaysia Tests AI in Judicial System." *Daily Sabah*, April 12, 2022, sec. Life. https://www.dailysabah.com/life/mr-robot-takes-on-laworder-malaysia-tests-ai-in-judicial-system/news.

Scott, Matthew. "Projected Growth of Remand Prison Population 'Alarming.'" *Newsroom*, April 4, 2022. https://www.newsroom.co.nz/page/projected-growth-of-remand-prison-population-alarming.

Lisa Owen *Judges' sentencing lacks consistency - Andrew Little* Newshub *2017-12-02 https://www.newshub.co.nz/home/politics/2017/12/judges-sentencing-lacks-consistency-andrew-little.html Accessed 2021-12-10*